\PassOptionsToPackage{dvipsnames,table}{xcolor}
\documentclass{opt2025_style/opt2025}

\usepackage[utf8]{inputenc}
\usepackage[T1]{fontenc}
\usepackage{lmodern}
\usepackage{microtype}
\usepackage{geometry}
\usepackage{natbib}
\usepackage{booktabs}
\usepackage{graphicx}
\usepackage{caption}
\usepackage{subcaption}
\usepackage{wrapfig}
\usepackage{multirow}
\usepackage{hyperref}
\hypersetup{colorlinks=true,linkcolor=MidnightBlue,citecolor=MidnightBlue,urlcolor=MidnightBlue}
\usepackage{url}
\usepackage{algorithm}
\usepackage{algorithmic}
\usepackage{enumitem}
\usepackage{minted}

\usepackage[capitalize]{cleveref}
\crefname{equation}{}{}  
\def\*#1{\mathbf{#1}}
\newcommand{\RR}{\mathbb{R}}
\usepackage{xspace}
\DeclareRobustCommand{\oplora}{\textsc{OPLoRA}\xspace}
\DeclareRobustCommand{\svdlora}{\textsc{SVDLoRA}\xspace}
\DeclareRobustCommand{\lorsum}{\textsc{LoRSum}\xspace}

\usepackage[colorinlistoftodos,textsize=scriptsize]{todonotes}

\title{Faster Than SVD, Smarter Than SGD: \\ The OPLoRA Alternating Update}
\optauthor{%
    \Name{Abdulla Jasem Almansoori}\textsuperscript{1} \Email{abdulla.almansoori@mbzuai.ac.ae} \\
    \Name{Maria Ivanova}\textsuperscript{2} \Email{ivanova.m.pe@gmail.com} \\
    \Name{Andrey Veprikov}\textsuperscript{1} \Email{andrei.veprikov@mbzuai.ac.ae} \\
    \Name{Aleksandr Beznosikov}\textsuperscript{3} \Email{anbeznosikov@gmail.com} \\
    \Name{Samuel Horv\'{a}th}\textsuperscript{1} \Email{samuel.horvath@mbzuai.ac.ae} \\
    \Name{Martin Tak\'{a}\v{c}}\textsuperscript{1} \Email{martin.takac@mbzuai.ac.ae} \\
    \addr
    \textsuperscript{1}MBZUAI, Abu Dhabi, UAE \\
    \textsuperscript{2}Yandex School of Data Analysis, Moscow, Russia \\
    \textsuperscript{3}Moscow Center for Advanced Studies, Moscow, Russia \\
}

\begin{document}
\maketitle

\begin{abstract}
Low-Rank Adaptation (LoRA) fine-tunes large models by learning low-rank updates on top of frozen weights, dramatically reducing trainable parameters and memory \citep{hu2022lora}. However, there is still a gap between full training with low-rank projections (\svdlora) and LoRA fine-tuning, indicating that LoRA steps can be further improved. In this study, we propose \oplora, a memory-efficient optimizer that closes this gap by casting LoRA optimization as an interpretable sub-problem and solving it efficiently with alternating least squares updates, where 1-2 alternating steps are empirically found to be sufficient to closely match truncated SVD without ever forming the full matrix.
We also retrieve the recently proposed preconditioning methods for LoRA \citep{zhang2024riemannian,zhang2023preconditioned,zhang2021preconditioned,tong2021accelerating} as a special case.
\oplora supports momentum by maintaining a low-rank estimate using the same subroutine (\lorsum) for computing the step, with a memory budget of 3 times the number of LoRA parameters (i.e., same as Adam).
We also propose an experimental scaled variant that uses the K-FAC metric \citep{martens2020optimizingneuralnetworkskroneckerfactored}, which could be of interest.
Across a linear task, MNIST, CIFAR-100, and RoBERTa-base (MNLI), \oplora consistently approaches \svdlora's performance using significantly less memory.\footnote{
    Code:
    \url{https://github.com/zeligism/OPLoRA}.
}
\end{abstract}

\section{Introduction}
\label{sec:intro}
Fine-tuning large pretrained models remains costly: updating all $N$ parameters demands substantial GPU memory and state tracking. Parameter-efficient methods -- especially \emph{LoRA} -- replace full fine-tuning with low-rank adapters $\*U\in\RR^{d_\text{out}\times r}$, $\*V\in\RR^{d_\text{in}\times r}$ added to a frozen base weight $\*W_0$ so that $\*W = \*W_0 + \*U\*V^\top$ with $r\ll\min(d_\text{out},d_\text{in})$ \citep{hu2022lora}.
Although LoRA reduces parameter and memory costs, first-order training of $(\*U,\*V)$ can be ill-conditioned \citep{tong2021accelerating} and sensitive to hyperparameters. Preconditioning with $\*V^\top\*V$ and $\*U^\top\*U$ remedies part of this \citep{zhang2024riemannian,zhang2023preconditioned,zhang2021preconditioned}, yet each step still acts only locally on the current factor subspaces.

We propose \textbf{Optimally Preconditioned LoRA (\oplora)}, which reframes the LoRA update as a small rank-$r$ least-squares/SVD subproblem solvable by \emph{alternating} updates up to arbitrary accuracy using a subroutine we call \lorsum.
\oplora retrieves previous preconditioned LoRA updates as special cases with one simultaneous step.
With two alternating steps and beyond, it starts to closely track the optimal rank-$r$ truncated SVD direction (\svdlora) up to arbitrary precision (see \cref{fig:linear}) without constructing, factorizing, or operating on a full matrix.
In this way, \oplora bridges vanilla LoRA training and the powerful yet costly SVD projection baseline while retaining LoRA’s parameter and memory efficiency.

\paragraph{Contributions.}
Our work makes the following contributions: 
(i) A unified view that generalizes preconditioned LoRA as a one-step instance of the \lorsum subroutine.
(ii) A practical algorithm (\cref{sec:method}) with $2 \times$ the memory for momentum (and $4 \times$ memory for the K-FAC-scaled version), avoiding full-matrix SVDs and dense operations. We empirically show that \oplora converges to \svdlora as the number of alternating updates increases (see \cref{fig:linear}).
(iii) Empirical gains over SGD/Adam LoRA on linear synthetic task, LeNet5 on MNIST \citep{mnist}, and positive results on RoBERTa \citep{liu2019robertarobustlyoptimizedbert} on GLUE \citep{wang2019gluemultitaskbenchmarkanalysis}. \oplora with 1-2 steps is competitive with \svdlora under similar budgets to LoRA + Adam with slightly more compute.

\section{Related Work}
\label{sec:related}
\textbf{Preconditioned LoRA.}
ScaledGD, PrecGD, and Riemannian preconditioning improve low-rank optimization by preconditioning LoRA updates \citep{tong2021accelerating,zhang2021preconditioned,zhang2023preconditioned,xu2023power,zhang2024riemannian}.
Our one-step \oplora recovers these methods, while multi-step alternating steps moves toward the best rank-$r$ direction.
Momentum with LoRA preconditioning has been explored in \citep{tastan2025loftlowrankadaptationbehaves}, but their projection method restricts momentum to the joint subspaces of all iterates, limiting its benefits \textbf{as shown in our experiments} (named \oplora Proj. in \cref{fig:linear}).
A recent work \citep{mahdavinia2025lowrankmomentumfactorizationmemory} proposes an efficient algorithm for updating a low-rank factorization of the momentum matrix, which is relevant to our work but differs substantially in implementation.

\textbf{Extensions of LoRA.}
Several recent works have extended the LoRA framework to improve training efficiency and generalization. 
\emph{AdaLoRA} \citep{zhang2023adalora} employs SVD decomposition
and prunes less significant singular values for more efficient
updates.
\emph{ReLoRA} \citep{lialin2023relora} periodically merges low-rank adapters into the base weights and resets new ones, allowing the model to gradually accumulate higher-rank updates.
\emph{DoRA} \citep{liu2024dora} decomposes pre-trained model weights into magnitude and direction, fine-tuning both: the former captures global scaling information, while the latter carries most representational capacity.
\emph{GaLore} \citep{zhao2024galore} proposes projecting the full gradient into a compact rank-$r$ subspace, updating in the small space, and projecting the update back to the full space.

Our method differs from prior literature by focusing on the preconditioning of LoRA such that they follow \svdlora as close as possible, bridging the gap between LoRA fine-tuning and full-tuning.
We note that exact and efficient methods for additive modification of a SVD exist \citep{BRAND200620} with overall complexity $\mathcal{O}(d_x d_y r)$.
While our approach is only approximate, it uses only thin operations with complexity $\mathcal{O}(K\max\{d_x, d_y\} r^2)$, where $K \geq 1$ is a hyperparameter that controls the accuracy of the approximation (see \cref{app:algo}), making it more flexible and particularly light under practical settings $K \in \{1, 2\}$.
In addition, its connection to the other methods in the literature can be of interest in itself.


\section{Method: Alternating Low-Rank Subproblems for LoRA}
\label{sec:method}
\paragraph{Setup and notations.} For a single layer with frozen $\*W_0\in\RR^{d_\text{out} \times d_\text{in}}$ we learn rank-$r$ adapters $\*U,\*V$ so that $\*W = \*W_0+\*U\*V^\top$.
Let $f(\*W)$ be the objective function and denote the full weight gradient by $\*G := \nabla_{\*W} f (\*W)$, where the other parameters, if any, are implicit.
LoRA gradients satisfy $\*G_\*U := \nabla_{\*U} f (\*W) = \*G\*V$ and $\*G_\*V := \nabla_{\*V} f (\*W) = \*G^\top \*U$.
Furthermore, for deep neural networks, let $B$ be the batch size, and let $\*X \in \RR^{B \times d_\text{in}}$ be the input to the linear layer $\*W$ and $\*S \in \RR^{B \times d_\text{out}}$ be the gradient of $f$ with respect to its output. Then, $\nabla_{\*W} f (\*W) = \*S^\top \*X$.

\paragraph{LoRA update sub-problem.}
Consider the regularized low-rank sub-problem approximating the next full-rank step $\*W_{t+1}$:
\begin{equation}
    \label{eq:subproblem}
    \min_{\*U,\*V}
    \quad
    \frac{1}{2} \left\|
        \*U \*V^\top - \*W_{t+1}
    \right\|_\text{F}^2
    + \mathcal{R}(\*U, \*V).
\end{equation}
The solution in terms of $\*U \*V^\top$ has a closed form when $\mathcal{R}(\*U, \*V) = 0$ under the Eckart-Young-Mirsky theorem, which is the rank-$r$ truncated SVD of $\*W_{t+1}$.
This is the ideal ``\svdlora'' scenario that we would like to replicate.
The solution in terms of $(\*U, \*V)$ is not unique since it can be written as $(\*U \*A, \*V\*A^{-1})$ for any invertible $\*A$.
Note the regularizer
$
\mathcal{R}_\text{fro}(\*U, \*V)
= \frac{\lambda_\*U}{2} \| \*U \|_\text{F}^2
+ \frac{\lambda_\*V}{2} \| \*V \|_\text{F}^2
$
with
$\lambda_\*U = \lambda_\*V > 0$,
gives $(\*U^*, \*V^*) = (\*U_r \*\Sigma_r^{1/2}, \*V_r \*\Sigma_r^{1/2})$,
where $\*U_r \*\Sigma_r \*V_r^\top$ is the rank-$r$ truncated SVD of $\*W_{t+1}$.
Equivalently, \svdlora updates its adapter to $\*U_r \*\Sigma_r \*V_r^\top$.

Assume an optimal solution exists\footnote{We can ensure its existence with a strongly convex $\mathcal{R}$.} and denote it as $(\*U_{t+1}, \*V_{t+1})$.
Consider the first-order stationary solutions of $\*U$ and $\*V$ given the optimal solution of the other matrix under $\mathcal{R}_\text{fro}$
\begin{align}
    \*U_{t+1}
    &\gets \*W_{t+1} \*V_{t+1} \left( \*V_{t+1}^\top\*V_{t+1} + \lambda_\*V \*I \right)^{-1},
    \label{eq:oplora-ideal-update-u}
    \\ \quad
    \*V_{t+1}
    &\gets \*W_{t+1}^\top \*U_{t+1} \left( \*U_{t+1}^\top \*U_{t+1} + \lambda_\*U \*I \right)^{-1}.
    \label{eq:oplora-ideal-update-v}
\end{align}
Note that it is impossible to directly compute the above updates due to the circular dependency between $\*U_{t+1}$ and $\*V_{t+1}$.
This expression suggests an alternating update strategy that uses the best solution so far for $\*U_{t+1}$ and $\*V_{t+1}$ each step, starting with some estimate.

Indeed, let $\*W_{t+1} := \*U_t \*V_t^\top - \eta \*G_t$,
and let $\*U_{t+1}^{(k)}$ and $\*V_{t+1}^{(k)}$ be the estimates for $\*U_{t+1}$ and $\*V_{t+1}$ at the alternating iteration $k$, where $\*U_{t+1}^{(0)} := \*U_t$ and $\*V_{t+1}^{(0)} := \*V_t$. Then,
\begin{align}
    \*U_{t+1}^{(k+1)}
    &\gets \left( \*U_t\*V_t^\top - \eta \*G_t \right) \*V_{t+1}^{(k)}
    \left(
        \left(\*V_{t+1}^{(k)}\right)^\top \*V_{t+1}^{(k)} + \lambda_\*V \*I
    \right)^{-1},
    \label{eq:oplora-alternating-update-u}
    \\ \quad
    \*V_{t+1}^{(k+1)}
    &\gets \left( \*V_t \*U_t^\top - \eta \*G_t^\top \right) \*U_{t+1}^{(k+1)}
    \left(
        \left(\*U_{t+1}^{(k+1)}\right)^\top\*U_{t+1}^{(k+1)} + \lambda_\*U \*I
    \right)^{-1}.
    \label{eq:oplora-alternating-update-v}
\end{align}
Observe how $\*V_{t+1}^{(k+1)}$ uses $\*U_{t+1}^{(k+1)}$ and not $\*U_{t+1}^{(k)}$.
Indeed, using a single simultaneous update for both $\*U$ and $\*V$ and recalling that $\*U_{t+1}^{(0)} = \*U_t$ and $\*V_{t+1}^{(0)} = \*V_t$, we get
\begin{align*}
    \*U_{t+1}
    &\gets \*U_t {\color{Gray} \*V_t^\top \*V_{t} (\*V_{t}^\top\*V_{t} + \lambda_\*V \*I)^{-1}}
     - \eta \*G_t \*V_{t} (\*V_{t}^\top\*V_{t} + \lambda_\*V \*I)^{-1},
    \\ \quad
    \*V_{t+1}
    &\gets \*V_t {\color{Gray} \*U_t^\top \*U_{t} (\*U_{t}^\top\*U_{t} + \lambda_\*U \*I)^{-1}}
     - \eta \*G_t^\top \*U_{t} (\*U_{t}^\top\*U_{t} + \lambda_\*U \*I)^{-1}.
\end{align*}
The gray terms cancel out as $\lambda_\*U, \lambda_\*V \to 0$.
The one-step simultaneous updates retrieves Riemannian preconditioned LoRA / ScaledGD / PrecGD \citep{zhang2024riemannian,tong2021accelerating,zhang2021preconditioned} and also gives a precise interpretation of the epsilon value in the inverse.

Our method is not a trivial extension to multiple alternating steps since we need to store $\*G_t$ and store it and reuse it efficiently, which requires hooks to save $\*S$ and $\*X$ in order to compute quantities like $\*G_t \*V_{t+1}^{(k)}$ (whereas a backward pass computes $\*G_t {\color{BrickRed} \*V_{t}}$, for example).
Note that the order of updating $\*U$ and $\*V$ is a custom choice and can affect performance depending on the conditioning of $\*U$ and $\*V$ (e.g., at LoRA initialization).
We adopt the order shown above.

\paragraph{Proximal updates.}
Using a proximal regularizer
$
\mathcal{R}_\text{prox}(\*U, \*V)
= \frac{\lambda_\*U}{2} \left\| \*U - \*U_t \right\|_\text{F}^2
+ \frac{\lambda_\*V}{2} \left\| \*V - \*V_t\right\|_\text{F}^2
$,
we can rewrite the subproblem in \cref{eq:subproblem} as
\begin{equation}
    \arg\min_{\*U, \*V} \
        \langle \*U \*V^\top, \*G \rangle_\text{F}
        + \tfrac{1}{2 \eta} \|
            {\color{BrickRed} \*U \*V^\top}
            -
            {\color{Brown} \*W_{t}}
        \|^2_\text{F}
        + \tfrac{\lambda_\*U}{2} \|
            {\color{OliveGreen} \*U}
            -
            {\color{RoyalBlue} \*U_t}
        \|^2_\text{F}
        {\color{Gray}
            + \tfrac{\lambda_\*V}{2} \|
                \*V
                -
                \*V_t
            \|^2_\text{F}
        }
        \label{eq:proximal-subproblem}
    .
\end{equation}
We ignore $\*V$ to keep things simple.
Let $\*W_t = \*U_t \*V_t^\top$ and solve for $\*U$ as in \cref{eq:oplora-ideal-update-u}
\begin{equation}
    \*U_{t+1}
    \gets
    \*U_t (
        {\color{Brown} \*V_t^\top \*V_{t+1} }
        +
        {\color{RoyalBlue} \lambda_\*U \eta \*I}
    ) (
        {\color{BrickRed} \*V_{t+1} \*V_{t+1}}
        +
        {\color{OliveGreen} \lambda_\*U \eta \*I}
    )^{-1}
    -
    \eta \*G_t \*V_{t+1} (
        {\color{BrickRed} \*V_{t+1}^\top \*V_{t+1}}
        +
        {\color{OliveGreen} \lambda_\*U \eta \*I}
    )^{-1}
\end{equation}
The colors show the dependency of the terms in the update and where they come from.
Note that $\*G_t$ can also be replaced with a momentum estimate $\*G_t + \alpha \mathcal{M}_{t-1}(\*G)$.
Further, note that a first-order approximation of the term on the left around $\eta \to 0$ gives an inverse-free update similar to a derivation of weight decay in AdamW (e.g., see \cite{zhuang2022understandingadamwproximalmethods}).
We invert explicitly since we need an inverse for the other terms.
The algorithms in \cref{app:algo} show the \oplora step with proximal regularization in detail.

\paragraph{Low-Rank Sum (\lorsum).}
In the above updates, we have assumed that $\*W_{t+1} := \*U_t \*V_t^\top - \eta \*G_t$ and set $\*U_{t+1}^{(0)} := \*U_t$ and $\*V_{t+1}^{(0)} := \*V_t$.
In general, we can approximate any arbitrary sum of low-rank\footnote{Not necessarily low-rank, but preferably.} matrices $\*W_{t+1} := \sum_{i=1}^n c_i \*U_{i,t} \*V_{i,t}^\top$ and a regularizer. Without loss of generality, we set $\*U_{t+1}^{(0)} := \*U_{1,t}$ and $\*V_{t+1}^{(0)} := \*V_{1,t}$, where $\*W_{t} = \*U_{1,t} \*V_{1,t}^\top$, for example, and consider a proximal regularizer on the factors to bound their updates.
We denote this operation as $\*U_{1,t+1} \*V_{1,t+1}^\top = \lorsum(\sum_{i=1}^n c_i \*U_{i,t} \*V_{i,t}^\top)$, where the initial estimate is implicitly $\*U_{1,t} \*V_{1,t}^\top$ (or whichever low-rank matrix being updated).
Thus, we can reuse the subroutine in \cref{eq:oplora-alternating-update-u,eq:oplora-alternating-update-v} to update the low-rank estimate of any matrix that is a sum of low-rank matrices without constructing it itself.

\paragraph{Subspace iteration.}
Let $\mathcal{P}_\*X := \*X (\*X^\top \*X)^{-1} \*X^\top$ be the projection onto the column space of $\*X$.
Consider \cref{eq:oplora-ideal-update-u,eq:oplora-ideal-update-v} with $\lambda_\*U = \lambda_\*V = 0$ for simplicity, and note
\begin{align*}
    \*U_{t+1} \*V_{t+1}^\top
    &\overset{(\ref{eq:oplora-ideal-update-u})}{=}
    \*W_{t+1} \*V_{t+1} (\*V_{t+1}^\top\*V_{t+1})^{-1} \*V_{t+1}^\top
    =
    \*W_{t+1} \mathcal{P}_{\*V_{t+1}}
    \\
    &\overset{(\ref{eq:oplora-ideal-update-v})}{=}
    \*U_{t+1} (\*U_{t+1}^\top\*U_{t+1})^{-1} \*U_{t+1}^\top \*W_{t+1}
    =
    \mathcal{P}_{\*U_{t+1}} \*W_{t+1}.
\end{align*}
Now consider \cref{eq:oplora-alternating-update-u,eq:oplora-alternating-update-v} at iteration $k$
\begin{equation}
    {\color{Brown} \*U_{t+1}^{(k+1)} } \left(\*V_{t+1}^{(k)}\right)^\top
    \overset{(\ref{eq:oplora-alternating-update-u})}{=}
    \*W_{t+1} \mathcal{P}_{\*V_{t+1}^{(k)}},
    \quad\quad\quad
    \*U_{t+1}^{(k+1)} \left({\color{Brown} \*V_{t+1}^{(k+1)} }\right)^\top
    \overset{(\ref{eq:oplora-alternating-update-v})}{=}
    \mathcal{P}_{\*U_{t+1}^{(k+1)}} \*W_{t+1}.
\end{equation}
This can be seen as a subspace iteration method on $\*W_{t+1}^\top \*W_{t+1}$ and $\*W_{t+1} \*W_{t+1}^\top$ \citep[Ch.\ 5]{subspaceiteration}.
We leave a detailed analysis of this for future work.

\paragraph{Momentum.}
Let $\mathcal{M}(\*G_\*U)$ and $\mathcal{M}(\*G_\*V)$ be the momentum buffers for the LoRA weights. Consider $\mathcal{M}(\*G_\*U)$ for brevity.
A naive implementation is to accumulate the preconditioned gradients
\begin{equation*}
    \mathcal{M}_{t}^\text{naive}(\*G_\*U)
    := \*G_{\*U, t} (\*V_t^\top \*V_t)^{-1} + \alpha \mathcal{M}_{t-1}^\text{naive}(\*G_\*U).
\end{equation*}
However, this would not preserve the property that the gradient lies in the subspaces of $\*U_t$ and $\*V_t$ since the momentum buffers contain $\*G_\tau \*U_\tau$ for $\tau < t$.
An intuitive fix is to re-project the momentum buffers before preconditioning them as done in LoFT~\citep{tastan2025loftlowrankadaptationbehaves}.
\begin{align*}
    \mathcal{M}_{t}^\text{proj}(\*G_\*U)
    &:= \*G_{\*U, t} (\*V_t^\top \*V_t)^{-1} + \alpha \mathcal{M}_{t-1}^\text{proj}(\*G_\*U) \*V_{t-1} \*V_t (\*V_t^\top \*V_t)^{-1}
    \\
    \implies
    {\color{Brown} \mathcal{M}_{t}^\text{proj}(\*G_\*U) \*V_t}
    &= (\*G_t + \alpha {\color{Brown} \mathcal{M}_{t-1}^\text{proj}(\*G_\*U) \*V_{t-1}} )\mathcal{P}_{\*V_t}
    ,
\end{align*}
However, this would make the momentum buffers be projected onto the joint subspace of all the iterations and would require saving the previous iterates.
It is not clear how to avoid this without having another buffer or a higher-rank LoRA.

In order to maintain the benefits in the subspace iteration section, we consider directly tracking a low-rank estimate of the full-rank momentum using the \lorsum subroutine in \cref{eq:oplora-alternating-update-u,eq:oplora-alternating-update-v} to sum the quantity $\*G_t + \alpha \mathcal{M}_{t-1}^\text{lor}(\*G)$, i.e.,
\begin{equation}
    \mathcal{M}_{t}^\text{lor} := \lorsum(\alpha \mathcal{M}_{t-1}^\text{lor}(\*G) + \*G_t),
\end{equation}
where $\mathcal{M}_{0}^\text{lor} = \*U_{\mathcal{M}(\*G),0} \*V_{\mathcal{M}(\*G),0}^\top = \*0$.
This is achieved by initializing $\*U_{\mathcal{M}(\*G),0}$ and $\*V_{\mathcal{M}(\*G),0}$ similarly to LoRA weights.
We compare the projected momentum version with ours in \cref{fig:linear} and show that our momentum buffer approaches (and outperforms) \svdlora with larger rank budgets.

\paragraph{Scaled \oplora}
We can endow the proximal terms in \cref{eq:proximal-subproblem} with non-Euclidean metrics to get a scaled version of \oplora. 
This method is still in the experimental phase, so we defer the details to \cref{app:scaled}. Note that the ``scale'' in ScaledGD \citep{tong2021accelerating} is LoRA preconditioning and not metric scaling as is the case here.

\section{Experiments}

\begin{figure}[t]
    \centering
    \includegraphics[width=0.35\linewidth]{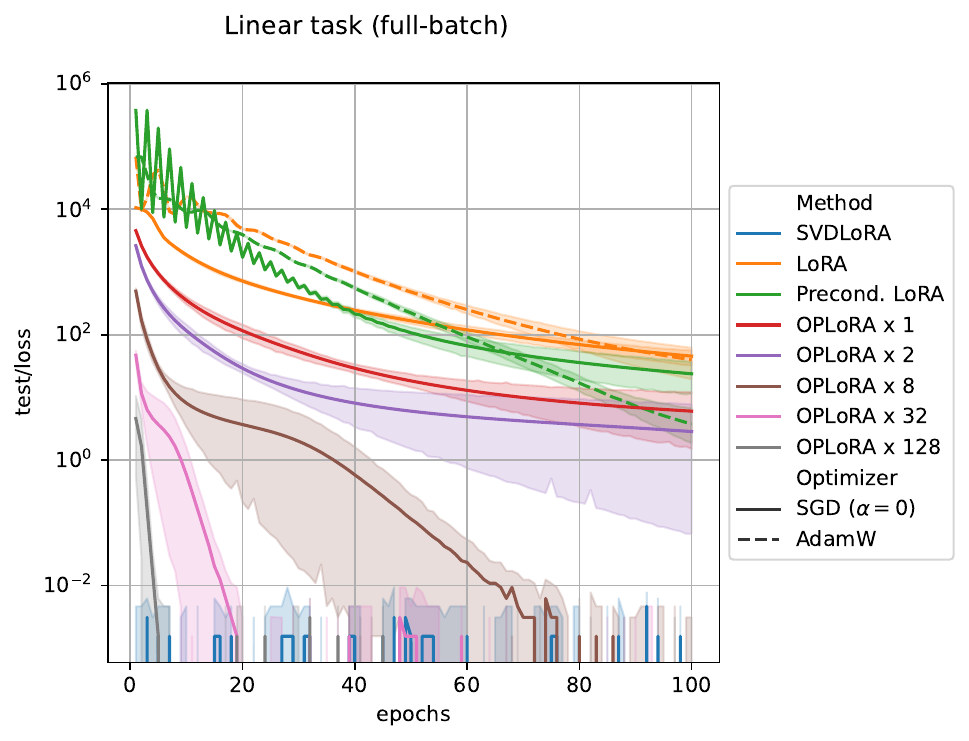}
    \includegraphics[width=0.36\linewidth]{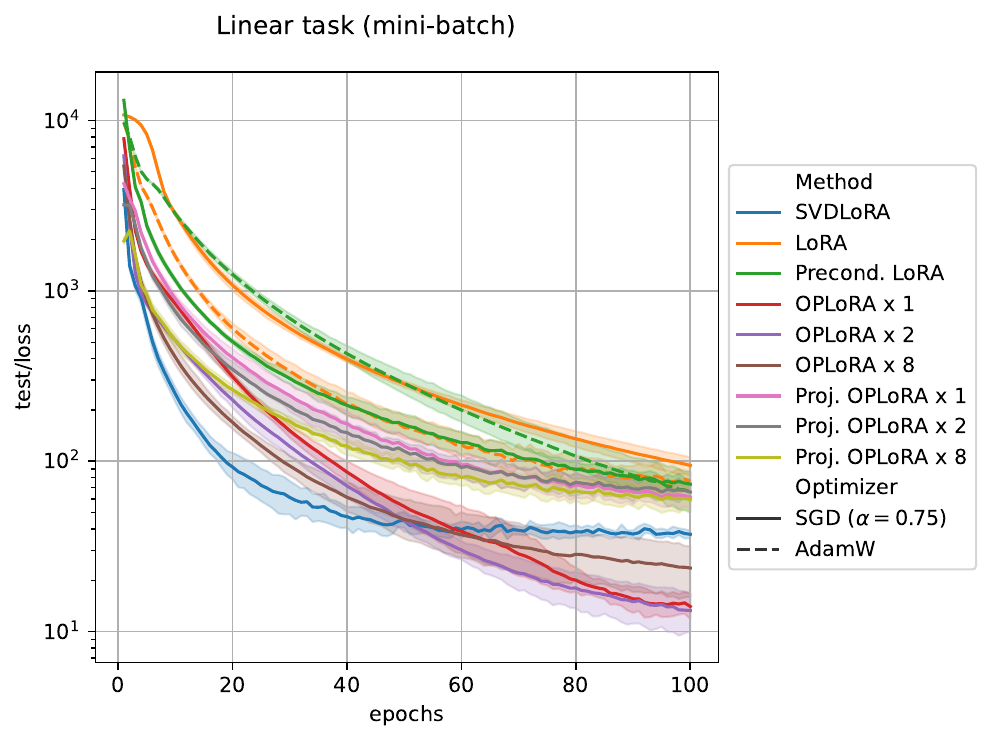}
    \includegraphics[width=0.27\linewidth]{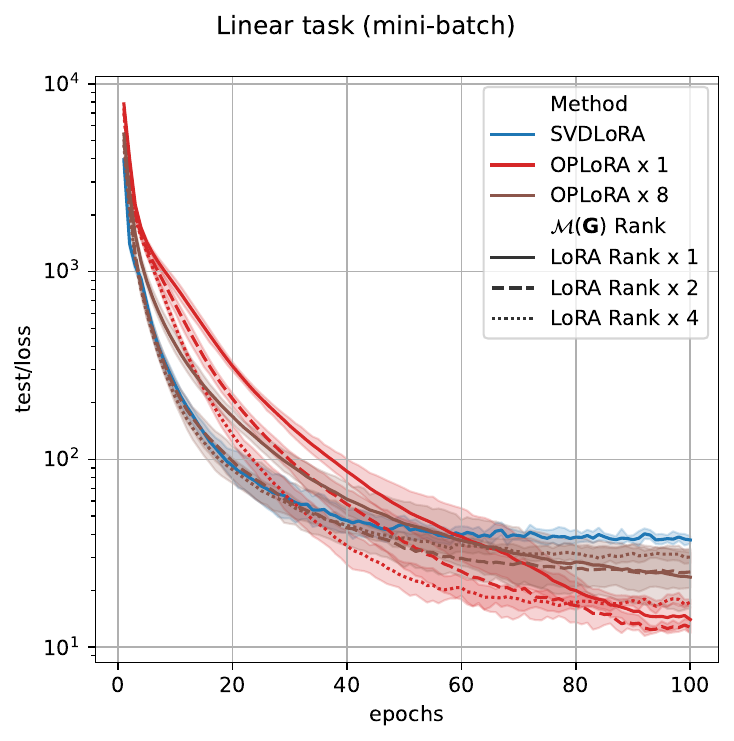}
    \caption{Linear task with full-batch (left) vs. mini-batch (middle) gradients.
    $\oplora \times k$ means $k$ alternating iterations.
    The right plot shows how \oplora approaches \svdlora as we increase the rank of $\mathcal{M}(\*G)$.
    Median runs with 95\% CI across 5 seeds are shown.}
    \label{fig:linear}
\end{figure}

\begin{wrapfigure}{r}{0.25\textwidth}
  \centering
  \vspace{-2em}
  \includegraphics[width=\linewidth]{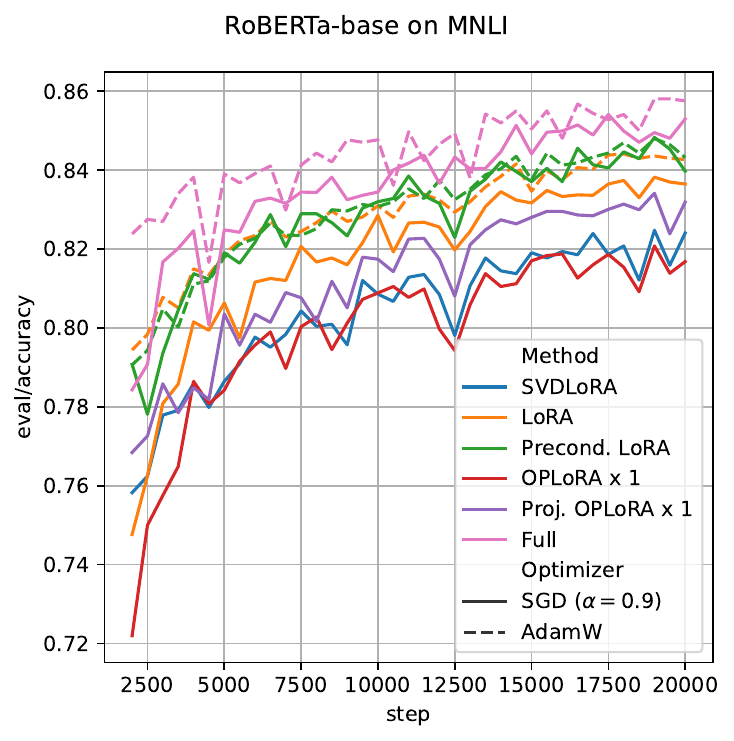}
  \caption{RoBERTa on MNLI task from GLUE.}
  \label{fig:glue}
\end{wrapfigure}

\label{sec:exp}
In all our experiments, linear layers\footnote{We still do not have an implementation for convolution layers, but it is fairly extensible by viewing the convolution layers as matrices.} are frozen and adapters are added.
\svdlora uses a full-rank adapter weight and does truncated SVD after every step, but note that all the optimization quantities, such as momentum, are still \emph{full-rank}.
For all experiments, we tune the learning rate of each method-optimizer variant to ensure a fair comparison.

\textbf{Baselines.}
We consider the following baselines: full weight  training / fine-tuning (upper bound), vanilla LoRA with SGD+momentum and AdamW, Preconditioned LoRA \citep{zhang2024riemannian,tong2021accelerating,zhang2021preconditioned}, \svdlora (full update then rank-$r$ projection via truncated SVD), \oplora, and \oplora with projected momentum \citep{tastan2025loftlowrankadaptationbehaves}.

\textbf{Linear task.}
The linear task is as simple as the subproblem \cref{eq:subproblem}.
We consider the objective $\min_{\*U, \*V} \| \*U \*V^\top - \*W \|_\text{F}$, where $\*W \in \mathbb{R}^{600 \times 200}$.
We consider full-batch gradients and mini-batch gradients by sampling the columns (i.e., batch size = 200 vs. 64).
The full-batch task uses SGD without momentum, whereas the stochastic task uses SGD with momentum.
Note that for the deterministic case, we expect \svdlora to converge in one step with $\eta = 0.5$, so we use the same learning rate for \oplora as well (for LoRA with SGD, we typically find the scaling $0.02 \eta$ to be optimal).
Adam always uses the default parameters with decoupled weight decay.
In \cref{fig:linear}, we show \oplora for a different number of alternating / lookahead iterations.
Observe how \oplora converges to \svdlora with more iterations.
The mini-batch task is more difficult despite its simplicity.
We tuned \svdlora and found that $\eta = 0.1$ and momentum $\alpha = 0.75$ yields the best result.
We use the same parameters for \oplora, which surprisingly outperforms \svdlora.
Interestingly, more alternating iterations approach \svdlora faster, which is good in the short horizon, but not necessarily good in the long horizon.
We do not yet have an explanation for this phenomenon.

\textbf{MNIST and CIFAR-100 task.}
In these two tasks, we use LeNet5 for MNIST and keep its convolution layers as they are and only add adapters to the linear classification layers.
For CIFAR-100, we use a custom MLP with one convolution layer that transforms the input into patches, and then we process the patches using linear layers throughout.
Non-LoRA layers are optimized using vanilla SGD with momentum.
These tasks are proofs of concept that demonstrate the feasibility of \oplora on more complex tasks without trying to achieve SOTA performance.
The exact architectures are not important but rather the performance of \oplora relative to \svdlora is, which is corroborated.
Results are shown in \cref{app:mnist}.

\textbf{RoBERTa-base on GLUE–MNLI task.}
We demonstrate the feasibility of \oplora on a real-world fine-tuning task.
We tweak a fine-tuning script from HuggingFace that uses RoBERTa \citep{liu2019robertarobustlyoptimizedbert} for sequence classification on GLUE \citep{wang2019gluemultitaskbenchmarkanalysis}.
In particular, we report results on the MNLI task in GLUE in \cref{fig:glue}.
We show only steps 2k to 20k for clarity.
Despite the performance, we believe that this result should be sufficient for proving the feasibility of \oplora that it indeed follows \svdlora.
It remains to understand why \svdlora does not perform well on this task and how to improve it.

\section{Discussion and Limitations}
\label{sec:discussion}

\oplora uses as much memory as Adam (i.e., 3 times the number LoRA parameters).
\oplora consists of thin multiplications as well as inversion of $r \times r$ matrices.
Still, its performance is not as good as LoRA optimization with Adam.
We believe that a multi-threaded implementation of \oplora can show significant gains in terms of wall-time performance, especially when the weight matrices are very large.
The \lorsum subroutine can start in parallel as soon as the backward pass reaches the layer because the input gradient (which is needed to continue the backward pass) does not depend on the preconditioned gradient.
Alternating steps can run for as long as the optimizer step function is not called, which can further improve the preconditioning of LoRA gradients.

\section{Conclusion}
\oplora casts LoRA optimization as an SVD subproblem and retrieves known algorithms, which makes the algorithm principled and interpretable.
Across diverse tasks, \oplora improves over vanilla LoRA and approaches truncated SVD (\svdlora) baselines under comparable budgets.
We hope this contributes a practical tool for memory-constrained fine-tuning of large models that scales with compute.


\bibliography{ref}

\clearpage
\appendix

\section{\oplora and \lorsum algorithms}
\label{app:algo}

\begin{algorithm}[H]
\caption{(Scaled) \oplora step}
\begin{algorithmic}[1]
    \STATE \textbf{Input:} $\*U,\*V$, $B$ (batch size), $\eta$ (learning rate), $\alpha$ (momentum), $\lambda_\*U = \lambda_\*V$ (factors weight decay), $\beta$ (scale smoothing), $\delta$ (scale damping).
    \STATE Save during forward: $\*X \in \mathbb{R}^{B \times d_x}$ (inputs).
    \STATE Save during backward: $\*S \in \mathbb{R}^{B \times d_y}$ (gradient w.r.t.\ outputs).
    \STATE \textit{\# Init or update scale}
    \IF{$\beta < 1$ and $\*D_\*U$ and $\*D_\*V$ not initialized}
        \STATE $\*D_\*U \gets \*I_{d_y}, \quad \*D_\*V \gets \*I_{d_x}$.
    \ELSE
        \STATE 
        $\*D_\*V \gets \lorsum(\beta \*D_\*V + (1 - \beta) \frac{1}{B} \*X^\top\*X),
        \quad
        \*D_\*U \gets \lorsum(\beta \*D_\*U + (1 - \beta) \frac{1}{B} \*S^\top\*S)$.
    \ENDIF
    \STATE \textit{\# Init momentum}
    \IF{$\alpha > 0$ and $\mathcal{M}(\*G)$ not initialized}
        \STATE $\mathcal{M}(\*G) \gets \*0$.
    \ENDIF
    \STATE \textit{\# \oplora step}
    \STATE $\*D_\*U^\delta := \*D_\*U + \delta \*I_{d_y}, \quad \*D_\*V^\delta := \*D_\*V + \delta \*I_{d_x}$
    \quad (damped scales)
    \STATE $\*U \*V^\top \gets \lorsum(\*U\*V^\top - \eta \*G - \eta \alpha  \mathcal{M}(\*G);\, \*D_\*U^\delta, \*D_\*V^\delta)$.
    \STATE \textit{\# Update momentum}
    \IF{$\alpha > 0$}
        \STATE $\mathcal{M}(\*G) \gets \lorsum(\alpha \mathcal{M}(\*G) + \*G)$.
    \ENDIF
\end{algorithmic}
\label{algo:oplora-step}
\end{algorithm}

\begin{algorithm}[H]
\caption{(Scaled) \lorsum with proximal regularization (\ref{eq:proximal-subproblem}) + (\ref{eq:scaled-norms})}
\begin{algorithmic}[1]
    \STATE \textbf{Input:} $\{c_i, \*U_i, \*V_i\}_{i=1}^{n}$, $\*D_\*U$, $\*D_\*V$, $K$.
    \quad ($\*U_1$ and $\*V_1$ are assumed to be the updated factors).
    \STATE \textbf{Usage:} $\lorsum(\sum_{i=1}^k c_i \*U_i \*V^\top;\, \*D_\*U=\*I_{d_y}, \*D_\*U = \*I_{d_x})$.
    \quad (Scales are identity by default.)
    \STATE $\*U_1^{(0)} \gets \*U_1, \quad \*V_1^{(0)} \gets \*V_1$.
    \FOR{$k = 0, \cdots, K-1$}
        \STATE $\*U_1^{(k+1)}
        \gets \left( c_1 \*U_1(\*V_1^\top \*V_1^{(k)} + \lambda_\*V \*I) + \sum_{i=2}^{n} c_i \*D_{\*U}^{-1} \*U_i\*V_i^\top \*V_1^{(k)} \right)
        \left( (\*V_1^{(k)})^\top \*D_{\*V} \*V_1^{(k)} \right)^{-1}
        $.
        \STATE $\*V_1^{(k+1)}
        \gets \left( c_1 \*V_1 (\*U_1^\top \*U_1^{(k)} + \lambda_\*U \*I) + \sum_{i=2}^{n} c_i \*D_{\*V}^{-1} \*V_i\*U_i^\top \*U_1^{(k)} \right)
        \left( (\*U_1^{(k)})^\top \*D_{\*U} \*U_1^{(k)} + \lambda_\*U \*I \right)^{-1}
        $.
    \ENDFOR
\end{algorithmic}
\label{algo:lorsum}
\end{algorithm}

We show in this section two algorithms, the (scaled) \oplora step function in \cref{algo:oplora-step}, which acts on LoRAs, and the (scaled) \lorsum with proximal regularization in \cref{algo:lorsum}, which is the main subroutine used in \oplora.
We show the scaled version as the regular version can be easily retrieved with identity scales (see \cref{app:scaled} for more information on the scaled version).

Note that the inputs and outputs of \lorsum are matrix products, but we sometimes write them implicitly as matrices to ease readability.
In general, they are low-rank matrices (except the gradient of the full weight, which is at most rank $B$, where $B$ is the batch size).
Furthermore, note that all the matrix operations in \lorsum are efficient in the sense that they are in the order of $\mathcal{O}(\max\{d_x, d_y, r\} r^2)$.

We attach a snippet of the source code of \lorsum in \cref{lst:lorsum} to demonstrate its simplicity and memory efficiency in practice.
The function \texttt{lorsum} sums factors\footnote{
    The format is confusing in hindsight and will be changed to $(\*U_i, \*V_i^\top)$ in the future.
} $(\*V_i^\top, \*U_i)$ with coefficients $c_i$ for $i = 1, \ldots, n$.
This code works as is, and the user can use it directly in their code.

An important implementation detail about the momentum buffer is that we use the full (unprojected) momentum step $\*W - \eta\*G - \eta \alpha \mathcal{M}(\*G)$ in the weight update (so that the momentum step potentially has higher rank) and then update the momentum buffer separately to get a low-rank estimate of $\*G + \alpha \mathcal{M}(\*G)$.

\begin{listing}[!ht]
\caption{Source code for the subroutine \lorsum in PyTorch.}
\label{lst:lorsum}
\begin{minted}[fontsize=\footnotesize, linenos, breaklines]{python}
def lorsum(
    factors: list[tuple[torch.Tensor, torch.Tensor]],
    coefficients: list[float],
    num_iters: int = 1,
    lmbd: float = 0.0,
    start_turn: str = "in",
) -> tuple[torch.Tensor, torch.Tensor]:
    assert len(factors) >= 2 and len(factors) == len(coefficients)
    solve = torch.linalg.solve

    # Pay attention to the _t suffix, these are the lookahead weights
    weight_in, weight_out = factors[0]
    weight_in_t, weight_out_t = weight_in.clone(), weight_out.clone()
    eye_r = torch.eye(weight_in_t.shape[0]).to(weight_in_t)

    for i in range(2 * num_iters):
        if start_turn == "in" and i % 2 == 0 or start_turn == "out" and i % 2 == 1:
            # ---------- Weight in ---------- #
            sum_in = weight_in.mul(lmbd)
            for coeff, (factor_in, factor_out) in zip(coefficients, factors):
                sum_in.add_((weight_out_t.T @ factor_out) @ factor_in, alpha=coeff)
            weight_in_t = solve(weight_out_t.T @ weight_out_t + eye_r.mul(lmbd), sum_in)
        else:
            # ---------- Weight out ---------- #
            sum_out = weight_out.mul(lmbd)
            for coeff, (factor_in, factor_out) in zip(coefficients, factors):
                sum_out.add_(factor_out @ (factor_in @ weight_in_t.T), alpha=coeff)
            weight_out_t = solve(weight_in_t @ weight_in_t.T + eye_r.mul(lmbd), sum_out.T).T

    return weight_in_t, weight_out_t
\end{minted}
\end{listing}

\section{Scaled \oplora}
\label{app:scaled}
We can endow the proximal terms in \cref{eq:proximal-subproblem} with non-Euclidean metrics to get a scaled version of \oplora. For example, consider the following norms induced by the metrics $\*D_\*U$ and $\*D_\*V$
\begin{equation}
    \| \*U \|_\text{U}^2
    =
    \langle \*U , \*D_\*U \*U \rangle_\text{F}
    ,\quad\quad
    \| \*V \|_\text{V}^2
    =
    \langle \*V , \*D_\*V \*V \rangle_\text{F}
    ,\quad\quad
    \| \*U \*V^\top \|_\text{UV}^2
    =
    \langle \*U \*V^\top , \*D_\*U \*U \*V^\top \*D_\*V \rangle_\text{F}
    .
    \label{eq:scaled-norms}
\end{equation}
These norms are consistent, i.e., it is easy to check that $\| \*U \*V^\top \|_\text{UV} \leq \| \*U \|_\text{U} \| \*V \|_\text{V}$.
We can use a K-FAC metric \citep{martens2020optimizingneuralnetworkskroneckerfactored} by maintaining (low-rank) running averages of $\frac{1}{B} \*X^\top \*X$ and $\frac{1}{B}\*S^\top \*S$ using the \lorsum subroutine\footnote{Damping is necessary to avoid non-positive definite metrics.}.
We can similarly use the Shampoo metric \citep{gupta2018shampoopreconditionedstochastictensor} to scale the LoRA gradients.
The derivation of the updates is a straightforward extension of the derivation from \cref{eq:proximal-subproblem}.
Note that we need to initialize the factors such that $\*D_\*U = \*I_{d_y}$ and $\*D_\*V = \*I_{d_x}$, which can be easily done by initializing the factors to be orthogonal to each other (e.g., take the QR of a random matrix and set the factors to $Q$).
Preliminary experiments on MNIST suggest that this method can minimize the training loss faster than all other methods, even with a low-rank estimate of the metrics $\*D_\*U$ and $\*D_\*V$.
We leave a proper treatment of the scaled \oplora for future work.

\begin{table}[htbp]
\centering
\caption{MNIST (LeNet5, batch size = 64) and CIFAR\textendash100 (PatchMLP, batch size = 64): mean test accuracy over time by method and optimizer. Higher is better. Bold is best, excluding full training.}
\label{tab:mnist}
\begin{tabular}{lccc}
\toprule
\multirow{2}{*}{\textbf{Method}}
& \multirow{2}{*}{\textbf{Optimizer}}
& \textbf{MNIST}
& \textbf{CIFAR\textendash100 } \\
&& \textbf{  Avg.\ Acc. (\%)}
& \textbf{  Avg.\ Acc. (\%)} \\
\midrule
LoRA           & SGD & $98.27 \pm 0.069$& $24.66 \pm 0.254$\\
LoRA           & AdamW  &$98.18 \pm 0.088$ & $27.16 \pm 0.116$ \\
Precond.\ LoRA & SGD  &$98.55 \pm 0.144$ &$27.53 \pm 0.360$ \\
Precond.\ LoRA & AdamW  &$98.32 \pm 0.056$ &$28.15 \pm 0.477$ \\
Proj. OPLoRA $\times$ 1 & SGD  &$98.53 \pm 0.098$ & $28.26 \pm 0.104$ \\
Proj. OPLoRA $\times$ 2 & SGD  &$98.51 \pm 0.062 $ & $\mathbf{28.29 \pm 0.061}$ \\
Proj. OPLoRA $\times$ 8 & SGD & $98.54 \pm 0.118$& $28.23 \pm 0.201$ \\
\midrule
OPLoRA $\times$ 1 & SGD  &$\mathbf{98.58 \pm 0.106}$ & $26.24 \pm 0.173$ \\
OPLoRA $\times$ 2 & SGD  &$98.55 \pm 0.100 $ & $26.47 \pm 0.178$ \\
OPLoRA $\times$ 8 & SGD & $98.56 \pm 0.134$& $26.94 \pm 0.190$ \\
\midrule
SVDLoRA        & SGD  &$ 98.52 \pm 0.074 $ &$27.48 \pm 0.365  $ \\
Full (upper bound)  & SGD & $98.77 \pm 0.111$ & $32.24 \pm 0.398$ \\
\bottomrule
\end{tabular}
\end{table}

\section{MNIST and CIFAR-100 experiments}
\label{app:mnist}
In \cref{tab:mnist}, we report the average accuracy over time over 3 seeds for both the MNIST and CIFAR-10 tasks.
Note that this metric is not the last iterate accuracy.
We measure the average accuracy over time to take into account early convergence, which is one of the strengths of \oplora.
For MNIST, we use LeNet5, and for CIFAR-100, we use a custom model we call PatchMLP, which ``patchifies'' the image into tokens using a single convolution layer and then runs the tokens through linear layers with a position embedding.
This model demonstrates that \oplora works well on non-trivial architectures.

\section{Experimental details}
\label{app:exp}
For all experiments, we tune the learning rates $\eta$ for each method across values $\{1, 2, 5\} \times 10^{-k}$, which is consistent with values used in practice.
The regularization constants are usually set to a default value of $\lambda_U = \lambda_V = 10^{-3}$, except for RoBERTa, where it is set to $10^{-2}$.
LoRAs for all tasks have rank $r=8$, except for CIFAR-100 where it is $r=16$.
LoRAs are often multiplied by a constant called alpha and divided by the rank.
We do not use such scaling, so the effective alpha is the rank.
For the linear task, we intitialize the LoRAs to the rank-$r$ truncated SVD of the full weight and set the full weights to zero.

We point the interested readers to look at the configuration files in the source code and the scripts that reproduce the experiments to see the exact details in an easily accessible and readable format.

\end{document}